# MDS-GNN: A Mutual Dual-Stream Graph Neural Network on Graphs with Incomplete Features and Structure

Peng Yuan and Peng Tang

*Abstract*—Graph Neural Networks (GNNs) have emerged as powerful tools for analyzing and learning representations from graph-structured data. A crucial prerequisite for the outstanding performance of GNNs is the availability of complete graph information, i.e., node features and graph structure, which is frequently unmet in real-world scenarios since graphs are often incomplete due to various uncontrollable factors. Existing approaches only focus on dealing with either incomplete features or incomplete structure, which leads to performance loss inevitably. To address this issue, this study proposes a mutual dual-stream graph neural network (MDS-GNN), which implements a mutual benefit learning between features and structure. Its main ideas are as follows: a) reconstructing the missing node features based on the initial incomplete graph structure; b) generating an augmented global graph based on the reconstructed node features, and propagating the incomplete node features on this global graph; and c) utilizing contrastive learning to make the dual-stream process mutually benefit from each other. Extensive experiments on six real-world datasets demonstrate the effectiveness of our proposed MDS-GNN on incomplete graphs.

## I. INTRODUCTION

In machine learning and data science [1-4, 25, 28, 29], graph representation learning has emerged as a pivotal field, aiming to transform complex graph-structured data into meaningful low-dimensional representations [5-7, 17-19, 73, 77]. The burgeoning interest in this domain has spawned a wide variety of methodologies tailored to different types of graph data and application scenarios [9-11, 13-15, 50, 60, 64, 85].

Graph Neural Networks (GNNs), as an important and novel branch of graph representation learning, have received extensive research attention in recent years [16-23]. It is suitable for various graph data analysis tasks such as node classification [24, 27, 68], link prediction [30-34], and graph clustering [35-39, 41]. Graph neural networks have shown unprecedented progress in the challenges of many real-world applications, for instance, traffic prediction [42-44], recommender system [45-47], anomaly detection [48, 49] and drug discovery [51, 52].

The success of GNNs on graph related problems in different domains is attributed to their effective message passing mechanism, which exploits the topology of the graph to iteratively aggregate information from neighbor nodes, such as GCN [16], GAT [53], GraphSAGE [12], and APPNP [54]. Therefore, the excellent performance of existing GNNs on graph-structured data relies on a prerequisite, i.e., the acquired graph data is complete. However, in real-world scenarios, this condition is not always fulfilled [8, 24, 56-58, 66, 67, 69-71]. Due to resource limitations, privacy protection, or technical challenges in the data collection process [62, 83], graph data is usually incomplete with incomplete features (Fig. 1(a)) and incomplete structure (Fig. 1(b)). Worse is the case where both exist simultaneously (Fig. 1(c, d)). For instance, (1) in social networks, as people's awareness of digital privacy continues to increase, more and more users are reluctant to disclose their complete social relationships [55], (2) in traffic networks, data may be missing for certain road segments due to factors such as insufficient technological means or sensor failures, and (3) in bioinformatics, intermolecular interactions or protein networks may be limited by experimental conditions that make the failure to detect the complete information. When the original data are incomplete, the performance of the GNNs model will decrease significantly, which results in substantial challenges for GNNs.

To resolve the aforementioned problem, some special model designs have been introduced by researchers to improve the capability of GNNs when data are deficient. To handle incomplete features, most methods attempt to impute missing node attributes from available information. SAT [40] implements joint distribution modeling of structure and attributes through shared latent space assumptions and distribution matching technology to complete node attribute completion. AmGCL [61] effectively handles the problem of missing node attributes by combining feature precoding with Dirichlet energy minimization and self-supervised contrastive learning. PCFI [59] introduces a novel graph data missing imputation scheme based on pseudo-confidence of node features. For incomplete structures, a significant amount of research has been conducted on the topic of graph structure learning, which aims to jointly learn optimized graph structures and their corresponding node representations. Pro-GNN [63] trains the GNN model and refines the graph structure alternately by directly treating the adjacency matrix as a learnable parameter. Simp-GCN [65] improves the original graph structure by implementing a node similarity metric. D2PT [72] handles graph learning tasks with weak information by utilizing long-range information propagation and global graph encoding and employs contrasting prototype alignment algorithms to improve model efficiency. However, most of these approaches only address missing data from a single aspect. A more challenging task arises:

➢ P. Yuan is with the School of Computer Science and Technology, Chongqing University of Posts and Telecommunications, Chongqing 400065, China (e-mail: pyp654321@outlook.com).
➢ P. Tang is with the College of Computer and Information Science, Southwest University, Chongqing 400715, China (e-mail: tangpengcn@swu.edu.cn).

*RQ.* **Can an effective framework be designed to improve the performance of existing benchmark GNNs on incomplete graphs?**

To address this challenging problem, this paper aims to implement a contrastive learning framework assisted by graph autoencoders on incomplete graphs. Specifically, on the initial stream, we use the supervised signal of the data itself to reconstruct the properties of the graph through a graph autoencoder and employ them to construct an augmented graph. On the other stream, the original data uses personalized propagation on the augmented graph. The two streams jointly guide the training of the model, thereby reducing the impact of data deficiencies. The main contributions of this article are as follows:

- We make the first attempt to study the graph learning problem with both incomplete structures and incomplete features where partial nodes are missing all attributes.
- We propose a novel graph neural network framework, termed MDS-GNN, which reconstructs node features on the original graph and subsequently generates an augmented structural global graph. It facilitates dual-stream information propagation on two graphs while leveraging contrastive learning to mutually benefit the processes.

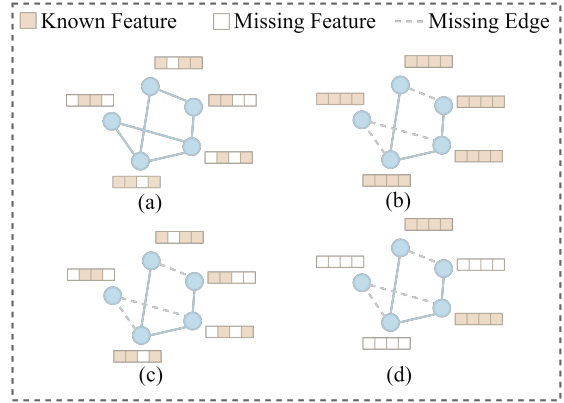

Figure 1. Illustrations of an incomplete graph.

## II. PRELIMINARIES

### A. Problem Formulation

We begin by introducing the relevant definitions and notations used in this paper.

***Definition 1:*** Given an undirected and attributed graph $G = (E, V) = (A, X)$, where $V = \{v_i | i = 1, 2, ..., n\}$ denotes a set of $n$ nodes, $E = \{e_{ij} | v_i, v_j \in V\}$ denotes a set of edges where nodes $v_i$ and $v_j$ are connected, $A \in \{0,1\}^{n \times n}$ is the binary adjacency matrix, where $A_{i,j} = 1$ if $e_{ij} \in E$ and 0 otherwise, and $X \in \mathbf{R}^{n \times f}$ is a matrix representing the set of all feature vectors $x_i \in \mathbf{R}^f$.

However, in real-world scenarios, the collected data are sometimes incomplete. Specifically, the feature matrix has some elements missing, or even the attributes of the entire node are missing, which is the case dealt with in this paper (Fig. 1(d)). Meanwhile, the structure of the graph can be incomplete, i.e., the data has only a limited set of edges. Based on the description mentioned above, incomplete graphs can be formulated by:

***Definition 2:*** Given a graph $G' = (V, E') = (X', A')$ with incomplete features and incomplete structure, where $E'$ is a subset of $E$ containing only some of the connectivity between nodes. $X' \in \mathbf{R}^{n \times f}$ denotes the incomplete feature matrix, and let $M = [m_1, m_2, ..., m_n]^T$ be the missing mask matrix, where $m_i$ is a d-dimensional all-0 or all-1 vector, which gives $X' = X \odot M$.

### B. Graph Autoencoders

According to some prior studies [74-76, 78], graph autoencoders aim to compress input data into a latent space and be able to reconstruct certain inputs from the obtained hidden representations. Concretely, an autoencoder typically consists of two components: encoder and decoder. The encoder is responsible for mapping the given information into a low-dimensional embedding, and the task of the decoder is to reconstruct the initial graph from the embedding under some supervised signals. They can be formalized by:

$$H = f_E(A, X), \tag{1a}$$

$$\tilde{G} = f_D(A, H), \tag{1b}$$

where $f_E(\cdot)$ denotes the graph encoder, $f_D(\cdot)$ denotes the graph decoder. $H \in \mathbf{R}^{n \times d_1}$ denotes the code encoded by the encoder and $\tilde{G}$ denotes the reconstructed graph.

### C. Contrastive Learning on Graphs

After significant successes in fields including computer vision and natural language processing, contrastive learning has garnered increasing attention from researchers aiming to extend its applicability to graph data. graph contrastive learning optimizes the encoder by maximizing the consistency between different streams of the same graph as the main objective. That is, the principle of maximizing mutual information (MI) is used to bring the representation of samples with similar semantic information closer and push the representation of unrelated samples away. It can be formulated as below:

$$\theta^* = \arg\min_{\theta} \mathcal{L}_{cl}(p(f_\theta(A_1, X_1), f_\theta(A_2, X_2))), \qquad (2)$$

where $p(\cdot)$ is a discriminator (e.g., cosine similarity), $A_1$ and $A_2$, $X_1$ and $X_2$ respectively represent the adjacency matrix and the node feature matrix under two streams.

## III. THE PROPOSED MDS-GNN MODEL

In this section, to address the poor performance of benchmark GNNs on incomplete graphs, a graph contrastive learning-based MDS-GNN model is proposed. Its framework is illustrated in Fig. 2, which consists of three parts:

- **The feature reconstruction module** based on parametric completion aims to assign parametric interpolation to missing node features, modeled by utilizing the superior reconstruction capabilities of GAE to extract valuable feature information from incomplete graphs.

- **The personalized propagation module** on augmented structure strives to utilize the augmented graph structure to augment incomplete initial graph information by personalized PageRank.

- **The node-level contrastive learning module** endeavors to maximize consistency between the original stream and the augmented stream, enabling the model to capture common knowledge from both streams.

### A. Feature Reconstruction Module

Given a graph G = (*A*, *X*) in **Definition 1**, GAE can target reconstructing feature X or structure A or both. Most previous GAEs are oriented towards link prediction and graph clustering tasks, so they usually chose to reconstruct *A*. In this module, we try to use GAE to reconstruct the features of the graph and use it for subsequent processing and model training.

According to the GAE form in Eq. (1), the encoder $f_E(\cdot)$ can usually be any type of GNN, e.g., GCN, GAT, etc. Since our encoder is based on incomplete graph, we choose the more expressive GAT as the encoder. In the node classification task, it can effectively learn the dependencies between nodes through the attention mechanism, thereby capturing more abundant and discriminative node features during the feature reconstruction process.

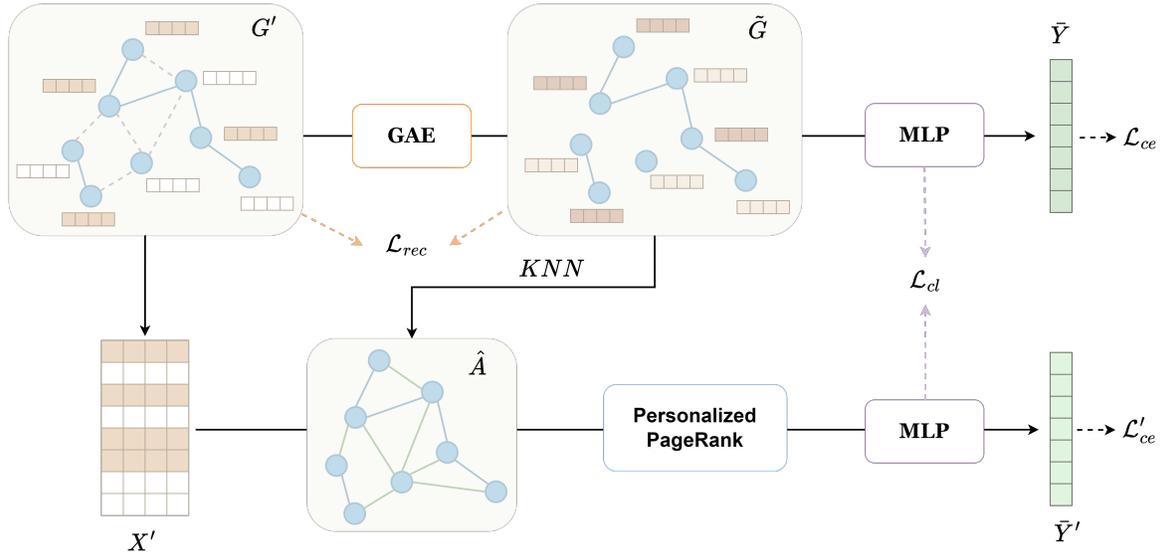

Figure 2. The overall framework of MDS-GNN.

For the set of nodes $\tilde{V} \subset V$ with missing features, we fill each of their features with a learnable vector $x_\omega \in \mathbf{R}^f$, that is, for $v_i \in \tilde{V}$, $x_i = x_\omega$, otherwise remains invariant. In addition, since we only use learnable parameters in the training stage, they will not appear in the inference stage. To narrow this gap, we choose to randomly replace the node features in $\tilde{V}$ with other known nodes with a smaller probability $p_r$. Our target is to use the existing structure $A'$ and feature $X'$ information to reconstruct the node representation $\tilde{X}$ of the graph through the graph autoencoder composed of GAT. The entire form can be simplified as:

$$\tilde{X} = f_{MLP}(f_{GAT}(A', X')), \qquad (3)$$

following the work of Veličković et al [53], $f_{GAT}(\cdot)$ is an encoder with multiple graph attention layers that maps node features to embeddings in a low-dimensional space. $f_{MLP}(\cdot)$ uses a simple MLP to decode the encoded latent embedding to obtain the reconstructed node representation.

As mentioned in **Definition 2**, since the input node feature $X'$ is incomplete, in order to enable GAE to reconstruct under the guidance of the correct signal, we introduce a feature reconstruction loss:

$$\mathcal{L}_{rec} = -\frac{1}{M}\sum_{i=1}^{M}\tilde{x}_i \ln x'_i, \tag{4}$$

where $M$ denotes the number of nodes with known attributes, $\tilde{x}_i$ and $x'_i$ denote the reconstructed and real features of node i.

### B. Personalized Propagation Module

Information propagation is the key operation in graph neural networks, which relies on the edges of the graph to complete information communication between adjacent nodes. Therefore, a complete graph structure plays a vital role in learning an accurate and comprehensive node representation. However, in our research problem, the edges of the graph are partially lacking, hindering the normal information propagation among nodes. To alleviate this problem, our motivation is to construct an augmented global graph and introduce a new stream of personalized propagation of the raw feature matrix on the global graph.

To be specific, we leverage the reconstructed complete feature matrix, employing k-Nearest Neighbors (kNN) graph as the augmented structure $\hat{A}$, which allows each node to have at least $k$ neighbors. The adjacency matrix $\hat{A}$ is expressed as:

$$\hat{A}_{ij} = \begin{cases} 1, & s_{ij} \geq \min(\varepsilon_{ik}, \varepsilon_{jk}), \\ 0, & \text{otherwise}, \end{cases} \tag{5}$$

where $s_{ij}$ is the cosine similarity between node vectors $\tilde{X}_i$ and $\tilde{X}_j$, $\varepsilon_{ik}$ denotes the similarity between $\tilde{X}_i$ and its $k$-th similar node vector in $\tilde{X}$, and $\hat{A}$ is essentially symmetrical.

Then, based on the approximated topic-sensitive PageRank [54], the personalized propagation process of node features is written as:

$$X^{(l+1)} = (1-\alpha)\tilde{\hat{A}}X^{(l)} + \alpha X', \tag{6a}$$

$$X^{(0)} = X', \quad \tilde{X}' = X^{(L)}, \tag{6b}$$

where $\tilde{\hat{A}} = \hat{D}^{-1/2}\hat{A}\hat{D}^{-1/2}$ is the symmetric normalized adjacency matrix, $\hat{D}$ denotes the degree matrix of $\hat{A}$, $\alpha \in (0,1]$ is the teleport probability, $L$ defines the number of power iteration steps, and $l \in [0, L-1]$, $\tilde{X}'$ is the propagated feature matrix.

### C. Node-level Contrastive Learning Module

After feature reconstruction and personalized propagation, the node representations $\tilde{X}$ and $\tilde{X}'$ of two streams are obtained. In the original stream, $\tilde{X}$ is mapped to the predicted label matrix $\bar{Y}$ for the classification task by an MLP-based transformation, and is formulated by:

$$\bar{Y} = softmax(\text{ReLU}(\tilde{X}W^{(0)})W^{(1)}), \tag{7}$$

where $W^{(0)}$ and $W^{(1)}$ are learnable weight matrices. After that, the cross-entropy error $\mathcal{L}_{ce}$ of all labeled samples is calculated based on $\bar{Y}$ for evaluation. Similarly, according to Eq. (7), we can also get the output $\bar{Y}'$ and loss function $\mathcal{L}'_{ce}$ in the contrastive stream. By joining the supervised signals of the two streams for training [80, 81], MDS-GNN is able to capture useful knowledge from both streams, consequently alleviating the effects of graph incompleteness.

However, on the original and augmented streams, due to the difference between the input structure $A'$ and $\hat{A}$, when using MLP with shared parameters to train on the two streams respectively, the representations generated by the same nodes may be significantly different. To bridge the gaps, we perform node-level contrastive learning to enhance the semantic consistency between two streams by maximizing the mutual information between them. To be specific, we map the MLP-encoded middle representations $H$ and $H'$ to another latent space using the linear projection layer, expressed as:

$$Z = HW^{(2)}, \quad Z' = H'W^{(2)}, \tag{8}$$

where $W^{(2)} \in R^{d_1 \times d_2}$ is a learnable weight matrix of the projection layer, and $Z, Z' \in \mathbf{R}^{n \times d_2}$ ($d_2$ is the embedding dimension) are the latent embeddings of the original and augmented streams, respectively.

Subsequently, in our framework, a contrastive loss $\mathcal{L}_{cl}$ derived from the normalized temperature-scaled cross-entropy loss (NT-Xent) [79] is applied:

$$\mathcal{L}_{cl} = -\frac{1}{2n}\sum_{i=1}^{n}\left(\log\frac{\varphi(z_i, z'_i)}{\sum_{k \neq i}^{n}\varphi(z_i, z'_k)} + \log\frac{\varphi(z_i, z'_i)}{\sum_{k \neq i}^{n}\varphi(z_k, z'_i)}\right), \tag{9}$$

where $\varphi(a,b) = e^{sim(a,b)/\tau}$, $sim(\cdot,\cdot)$ is the cosine similarity function, $\tau$ is the temperature parameter. $\mathcal{L}_{cl}$ increases the consistency of projections $z$ and $z'$ of the same node on two streams, which helps the model to collect knowledge common to both streams.

Finally, the losses of the various components are combined with the trade-off coefficients $\lambda$, $\mu$ and $\gamma$, so that the loss function for the whole framework is defined as follows:

$$\mathcal{L} = \mathcal{L}_{ce} + \lambda \mathcal{L}'_{ce} + \mu \mathcal{L}_{rec} + \gamma \mathcal{L}_{cl}. \tag{10}$$

## IV. Experiments

In this section, we conduct extensive experiments on the node classification task with various datasets to demonstrate the effectiveness of the proposed MDS-GNN model in dealing with incomplete graphs.

### A. Experimental Setups

*Datasets.* We evaluate on 6 public real-world datasets, including citation network [82], webpage network [83], Wikipedia network [85], and co-purchase graph [86]. The statistical and descriptive details of the dataset are shown in TABLE I. For the Amazon Photo dataset, we randomly select 20 nodes from each class for training, 30 nodes per class for validation, and the rest for testing. For rest datasets, we follow the public settings.

*Baselines.* To evaluate the effectiveness of MDS-GNN, we compare MDS-GNN with representative methods, including: (1) classic GNNs, including GCN [16], and GAT [53]; (2) GNNs with feature imputation, including PCFI [59]; (3) GNNs with graph structure learning, including SimP-GCN [63] and D2PT [72].

*Experimental Details.* We implement our model using PyTorch 1.12.1 and DGL 1.1.2. To acquire the objective results, all experiments are conducted on a GPU server with NVIDIA GeForce RTX 3050.

*a) Scenario implementation*: We perform random disturbances on the graph data to simulate the situation of incomplete graphs that occur in the real world. Specifically, to construct incomplete features, we randomly select 50% of the nodes in the graph and set all of the attributes of these nodes to unknown/missing values (i.e., 0). To construct incomplete structures, we drop 50% of the edges from the raw graph structure randomly.

*b) Hyper-parameter settings*: In our experiments, a grid search was conducted on the following hyperparameters. We search the learning rate in {0.001, 0.005, 0.01}, weight decay in {5e-4, 5e-3, 6e-3}, number of epochs in {200, 500, 800}, number of hidden units in {32, 64, 128, 256, 512}, replacement rate $p_r$ in {0, 0.05, 0.1}, input feature dropout in GAT in {0.2, 0.3}, number of hidden attention heads in GAT in {2, 4, 8}, number of hidden layers in GAT in {1, 2}, number of neighbors in KNN in {5, 10, 15, 20}, power iteration number $L$ in {5, 10, 15, 20}, weight coefficient $\lambda$ in {0.1 to 0.9}, weight coefficient $\mu$ in {0.1 to 0.9}, and weight coefficient $\gamma$ in {1, 1.5, 2, 4, 5}. Furthermore, we set the temperature $\tau$ in $\mathcal{L}_{cl}$ to 0.2, and set the teleport probability $\alpha$ to 0.01.

For our baselines, we use their official open-source codes and carefully tune some hyperparameters based on their default parameter settings to achieve the best performance. For all experiments, with the same dataset processing, we conduct 5 experiments under different seed settings and report their mean and standard deviation.

### B. Comparison Performance

The empirical analysis is started by comparing our proposed MDS-GNN method with traditional and state-of-the-art GNN models.

TABLE II shows the classification accuracy of our method with respect to the other baselines. As can be observed, (1) our proposed model achieves state-of-the-art performance on 5 of the 6 baseline datasets and second place results on the remaining dataset. Such competitive performance is attributed to a novel framework for self-information restoration and reinforcement guided by graph autoencoders and contrastive learning. (2) The performance of traditional GNN will drop significantly when the data is defective. And the methods that make improvements unilaterally bring limited improvement, and even weaker performance than traditional GNNs on some datasets. These results confirm the effectiveness of the proposed MDS-GNN framework, and more, they also show that besides relying on label information, rational use of the supervision signals of the data itself can also bring positive benefits to the model.

### C. Parameter and Ablation Studies

To provide a comprehensive evaluation of our model, we conducted parameter sensitivity analysis and ablation study on three datasets (Cora, CiteSeer, and Chameleon).

*Sensitivity analysis:* We study the sensitivity of critical parameters, including:

*a) Number of neighbors k:* We adjust $k$ in the range {5, 10, ...,40} to test the sensitivity. As shown in Fig. 3(a), in general, a smaller $k$ will have better performance. And too large may add some new noisy neighbors, which will degrade the performance.

*b) Effect of iteration number L:* To study the effect of the number of iteration steps $L$, we tune $L$ for the three datasets in

TABLE I. THE STATISTICAL INFORMATION OF DATASETS.

| Dataset | Nodes | Edges | Features | Classes | Train/Val/Test | Description |
|---|---|---|---|---|---|---|
| Cora | 2,708 | 10,556 | 1,433 | 7 | 140/500/1000 | citation network |
| CiteSeer | 3,327 | 9,228 | 3,703 | 6 | 120/500/1000 | citation network |
| Texas | 183 | 295 | 1,703 | 5 | 87/59/37 | webpage network |
| Wisconsin | 251 | 515 | 1,703 | 5 | 120/80/51 | webpage network |
| Chameleon | 2,277 | 36,101 | 2,325 | 5 | 1,092/729/456 | Wikipedia network |
| Amazon Photo | 7,650 | 238,163 | 745 | 8 | 160/240/7250 | co-purchase graph |

TABLE II. THE COMPARISON RESULTS OF NODE CLASSIFICATION ACCURACY (MEAN IN PERCENT ± STANDARD DEVIATION) IN INCOMPLETE GRAPHS.

| Methods | Cora | CiteSeer | Texas | Wisconsin | Chameleon | Amazon Photo |
|---|---|---|---|---|---|---|
| GCN | 60.52±2.30 | 48.50±1.08 | 57.30±2.02 | 45.88±6.86 | 51.23±1.63 | 85.64±1.88 |
| GAT | 62.02±1.88 | 48.52±1.75 | 57.84±2.76 | 47.06±5.41 | 46.62±2.34 | 86.60±1.90 |
| SimP-GCN | 64.73±2.11 | 50.44±1.69 | **70.27±3.82** | **60.78±4.47** | 49.30±1.35 | 86.25±0.71 |
| D2PT | **69.72±2.04** | **58.90±1.35** | 65.41±3.15 | 55.29±3.37 | 47.72±2.05 | 86.39±0.66 |
| PCFI | 67.42±1.95 | 56.24±0.92 | 65.95±3.67 | 53.33±2.29 | **51.89±1.93** | **86.73±1.24** |
| MDS-GNN(ours) | **70.12±0.90** | **58.42±0.53** | **73.51±2.02** | **61.57±2.00** | **56.01±1.76** | **88.53±1.24** |

a. the best and runner-up results are highlighted in red and blue respectively.

the range {5,10, ...,45}. As shown in Fig. 3(b), the optimal choice is different for each dataset, and we notice that any too-large $L$ leads to a significant performance degradation.

*Ablation study:* We illustrate the effect of feature reconstruction and dual-stream comparison by removing the corresponding losses. As shown in Fig. 4, it is evident that feature reconstruction plays a key role and the design of the comparison framework leads to notable improvements.

In addition, as shown in Fig. 5, we test our method on three datasets under different missing rates. Fig. 5(a) compares the performance of MDS-GNN under different missing rates for different classes of datasets, noting that the results at 0.4 are slightly better than 0.3, which we speculate is due to the randomness of the missingness. Fig. 5(b) compares the performance of using different backbone networks in the feature reconstruction module, and it can be observed that GAT outperforms GCN, and it is more obvious at high missing rates.

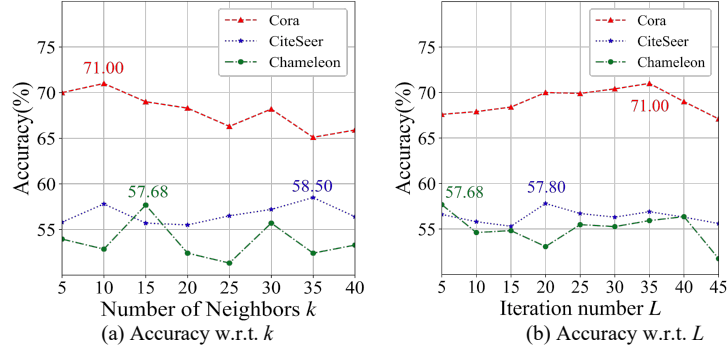

Figure 3. Sensitivity of hyper-parameters $k$ and $L$.

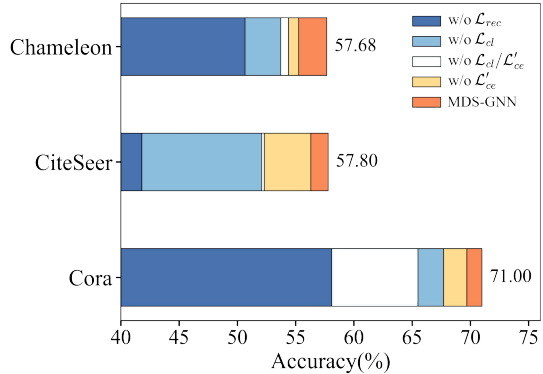

Figure 4. Ablation experiments on three datasets with 50% missing rate.

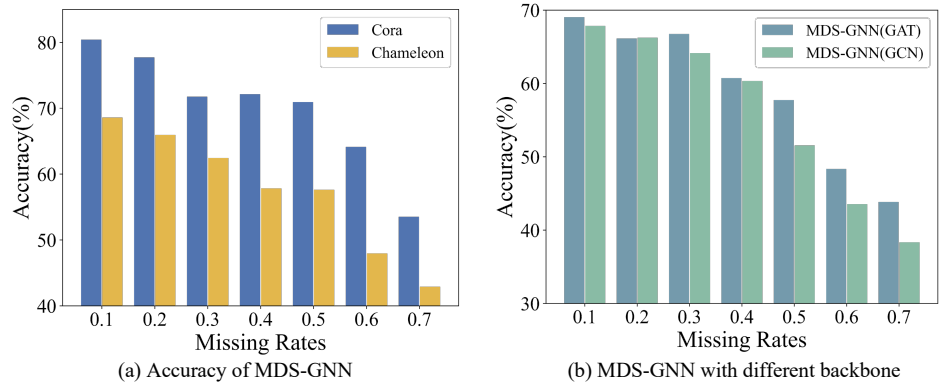

Figure 5. Comparisons of accuracy at different missing rates.

## V. CONCLUSION

In this paper, we attempt the challenging graph learning problem on graphs with incomplete structure and some nodes with all features missing. From the perspective of reconstructing features and augmenting the existing information by contrast stream, we propose a novel GNN model named MDS-GNN, which can effectively alleviate the problems caused by incomplete graphs. In the future, improvements can be made by: (1) trying to solve more downstream problems such as graph classification and link prediction, and (2) unsupervised graph learning that discards label information.